\RequirePackage{amsmath}   
\RequirePackage{amssymb}   

\documentclass[runningheads]{llncs}
\usepackage[T1]{fontenc}
\usepackage{graphicx}
\usepackage[colorlinks=true, linkcolor=blue, urlcolor=blue, citecolor=blue, bookmarks=false]{hyperref}
\usepackage{color}

\usepackage{enumitem}
\usepackage{multirow}
\usepackage{marvosym}

\begin{document}
%
\title{Confidence Calibration for Multimodal LLMs: \\ An Empirical Study through Medical VQA}

\author{Yuetian Du\inst{1} 
\and
Yucheng Wang\inst{2} 
\and
Ming Kong\inst{1} 
\and 
Tian Liang\inst{1} 
\and
Qiang Long\inst{3} 
\and 
Bingdi Chen\inst{3,4} 
\and 
Qiang Zhu\inst{1} \textsuperscript{(\Letter)}} 
\authorrunning{Y. Du et al.}

\institute{College of Computer Science and Technology, Zhejiang University \\
 \and
School of Computer Science and Technology, Xidian University\\
 \and
Zhihui Medical Technology (Shanghai) Co., Ltd., Shanghai, China \\
\and
The Institute for Biomedical Engineering \& Nano Science, Tongji University \\
\email{\{duyuetian2002, zjukongming, liangtian2022, zhuq\}@zju.edu.cn} \\
\email{wangyucheng2004@stu.xidian.edu.cn} \\
\email{alex.long@petctc.com} \\
\email{inanochen@tongji.edu.cn}}
\maketitle              
%

\begin{abstract}
Multimodal Large Language Models (MLLMs) show great potential in medical tasks, but their elicited confidence often misaligns with actual accuracy, potentially leading to misdiagnosis or overlooking correct advice. This study presents the first comprehensive analysis of the relationship between accuracy and confidence in medical MLLMs. It proposes a novel method that combines Multi-Strategy Fusion-Based Interrogation (MS-FBI) with auxiliary expert LLM assessment, aiming to improve confidence calibration in Medical Visual Question Answering (VQA). Experiments demonstrate that our method reduces the Expected Calibration Error (ECE) by an average of 40\%{} across three Medical VQA datasets, significantly enhancing MLLMs' reliability. The findings highlight the importance of domain-specific calibration for MLLMs in healthcare, offering a more trustworthy solution for AI-assisted diagnosis.


\end{abstract}
%

\section{Introduction}

Multimodal Large Language Models (MLLMs)~\cite{li2024llavamed,llavanext2024,liu2024improved,deitke2024molmo} have demonstrated significant application potential across various fields, due to their exceptional ability to integrate textual and visual information. However, these models commonly exhibit over-confidence~\cite{xiong2023uncertainty,liu2023uncertainty} in their predictions in practical applications, where there is a notable discrepancy between the confidence assigned to their predictions and their actual accuracy. This issue is particularly pronounced in high-stakes scenarios such as medical diagnosis. For instance, during clinical diagnosis, doctors typically rely on an iterative cycle of hypothesis generation, testing, and validation, integrating multi-source information such as patient history, laboratory results, and imaging data for comprehensive judgment. When MLLMs are introduced as decision-support tools, the confidence of their predictions must be precisely calibrated to ensure that doctors can effectively utilize the model's recommendations. Respectively, over-confidence in erroneous predictions can lead to misdiagnosis, while under-confidence may cause doctors to overlook correct advice. Therefore, calibrating the confidence of MLLMs in Medical Visual Question Answering (Medical VQA)~\cite{hasan2018overview,lau2018dataset,liu2021slake} tasks to ensure their reliability has become an urgent need in the practical application of AI-assisted diagnosis.

Although significant progress has been made in confidence calibration methods for Large Language Models (LLMs), the application of these methods in multimodal scenarios, especially in the medical field, is still in its infancy. The motivation for this study stems from the increasing prevalence of MLLMs in medical applications and the practical need for these models to provide accurate and reliable confidence estimates. By developing and evaluating confidence calibration methods specifically tailored for medical MLLMs, we aim to enhance the credibility of these models in clinical settings, thereby assisting healthcare professionals in making more accurate diagnostic decisions.

\noindent\textbf{Related Work.} In the field of confidence calibration for LLMs, there has been considerable research.~\cite{geng2023survey,wen2024abstention,liu2023trustworthy} Early methods primarily focused on using LLMs' token-likelihoods~\cite{andrey2020,kadavath2022,si2022reexamining,ahuja2022calibration} as confidence for calibration. However, as such white-box methods are not applicable to closed-source commercial models like ChatGPT, recent research has gradually shifted towards calibrating the verbalized confidence~\cite{xiong2023uncertainty,ni2024retrieval} of LLMs, which has been empirically shown to achieve better calibration effects~\cite{tian2023calibration} compared to token-likelihood confidence. Specific calibration methods include temperature scaling~\cite{si2022reexamining,wang2024verb,ahuja2022calibration}, prompting strategy~\cite{xiong2023uncertainty,ni2024retrieval,tian2023calibration,2025explain}, and reinforcement learning-based methods~\cite{kadavath2022,tao2024reinforce}. Additionally, self-consistency methods based on repeated sampling~\cite{savage2024medicalcalib} or multi-sampling at different temperatures~\cite{xiong2023uncertainty} have also proven effective in calibrating LLMs. Among these, prompting strategy has become the most widely used calibration method due to its plug-and-play nature and good portability. By designing a series of strategy templates, prompting strategy can effectively mitigate MLLMs' over-confidence issue, thus this study also adopts this method. It is noteworthy that in the medical field, limited related research~\cite{savage2024medicalcalib,wada2024} has mainly focused on the calibration of LLMs, while multimodal calibration, especially in Medical VQA, has not received sufficient attention. Although calibration strategies for LLMs can be transferred to multimodal scenarios, the effectiveness of these methods in the medical field has not been widely evaluated.

The main contributions of this study include the following three aspects:

\begin{itemize}[label=\textbullet]
    \item \textbf{Empirical Study:} We conducted the first comprehensive empirical study on the relationship between the accuracy of MLLMs and their self-assessed confidence in Medical VQA tasks, providing important insights into the calibration needs of multimodal LLMs in high-risk application scenarios.

    \item \textbf{New Calibration Method:} We proposed a new calibration method that combines a \textbf{M}ulti-\textbf{S}trategy \textbf{F}usion-\textbf{B}ased \textbf{I}nterrogation (MS-FBI) system with an auxiliary expert LLM assessment framework to better calibrate confidence scores with actual accuracy.

    \item \textbf{Experimental Validation:} We applied this method alongside various LLM baseline methods to a Medical VQA dataset, and the results show that our method significantly outperforms existing technologies, with an average improvement of 40\% in Expected Calibration Error (ECE), demonstrating the effectiveness of this method in real-world medical applications.
\end{itemize}

\begin{figure}
\includegraphics[width=\textwidth]{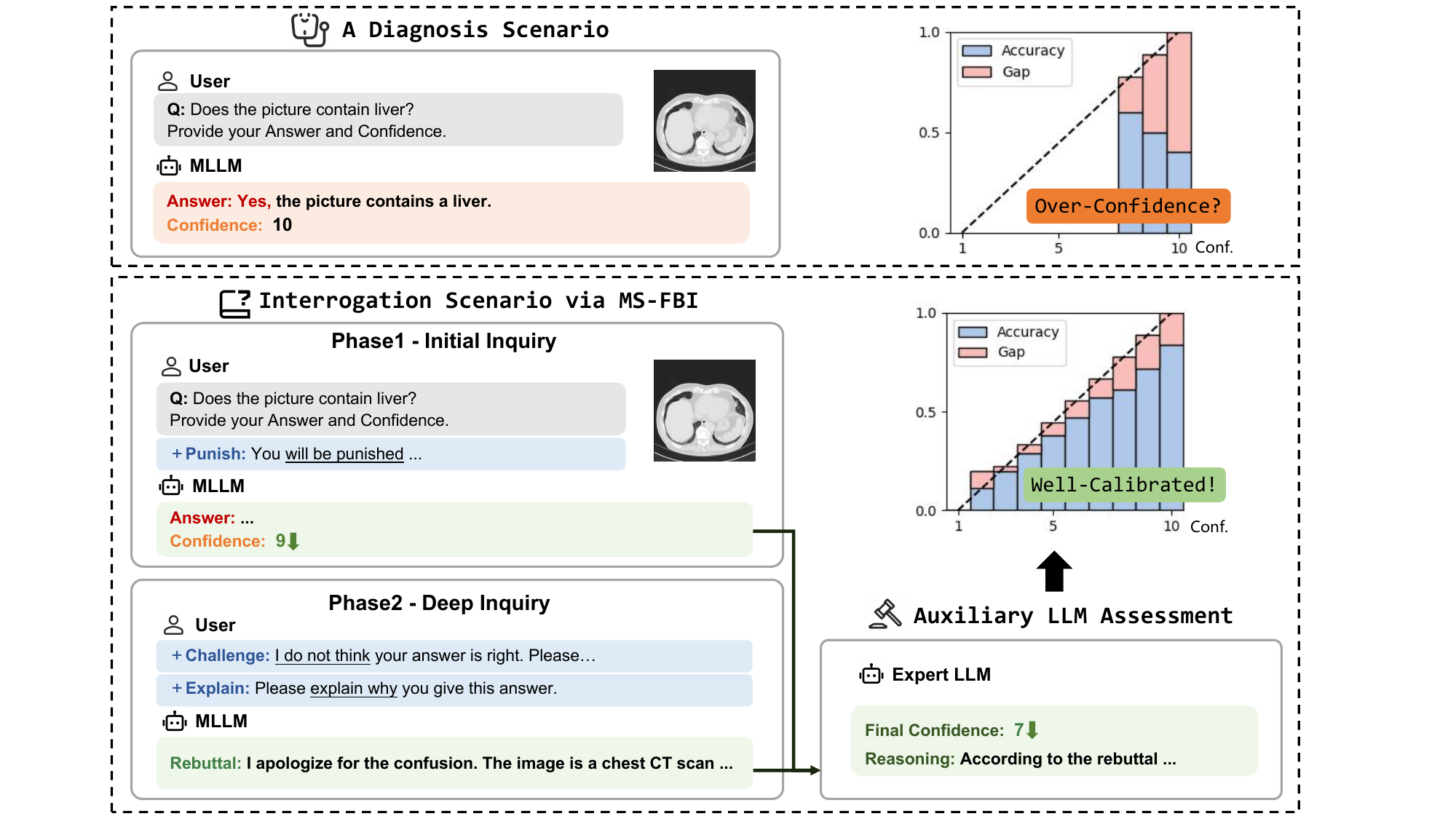}
\caption{The MLLM initially overconfidently identifies a liver in a chest CT scan. Through a two-phase interrogation process (MS-FBI), including an initial inquiry and deep inquiry with expert LLM assessment, the model's confidence is adjusted to a well-calibrated level.} \label{fig1}
\end{figure}

\section{Method}

\subsection{Overview}
As shown in Figure \ref{fig1}, the proposed method consists of two core components: an interrogation system based on a two-phase multi-strategy fusion approach (MS-FBI, down left), for collecting empirical information; and an auxiliary expert LLM assessment framework (down right), offering final judgement and analysis. This workflow not only achieves calibration between confidence and accuracy but also reveals MLLMs' psychological state under different interrogation scenarios. Through this systematic approach, we obtain auxiliary qualitative reasoning information about the MLLM's behavior, which provides potential opportunities to further enhance question-answering accuracy through the calibration process.

\subsection{Multi-Strategy Fusion-Based Interrogation (MS-FBI)}
Inspired by the lie detection process in criminal interrogations (FBI, e.g.)~\cite{manea2021lie}, this study designs a interrogation system that integrates multiple strategies based on prompt engineering, constructing a dynamically adaptive interrogation scenario, and systematically capturing the deep cognitive information of MLLMs.

\noindent\textbf{Initial Inquiry Phase—Punishment Mechanism.} In the first round of interaction, the system requires the MLLM to simultaneously output the answer $A$ and its confidence score $C$ for a Medical VQA question $Q$. This phase introduces a penalty constraint mechanism ($Punish$) by conditionally embedding the prompt "You will be punished if the answer is wrong but you answer it with high confidence." This strategy aims to curb MLLMs' overconfidence tendency and encourage more cautious and accurate answers. With sampled and comparable MLLM outputs, this design not only simulates the error cost mechanism in real-world decision-making but also effectively evaluates MLLMs' confidence calibration characteristics under pressure.

\noindent\textbf{Deep Inquiry Phase—Dual Verification Strategy.} In subsequent interactions, the system adopts a dual-track verification mechanism of $Challenge$ and $Explain$. By embedding prompts behind the context of the previous round of dialogue, the system obtains MLLMs' rebuttal $R_{mllm}$. Logical challenges expose contradictions in MLLMs' reasoning chain through targeted questioning, forcing it to review and correct its answers. This strategy draws on the core principles of cognitive restructuring. Meanwhile, the explanation reinforcement requires MLLMs to provide a step-by-step interpretation of its answer generation process, revealing potential knowledge gaps or logical flaws. The synergistic application of these two strategies significantly enhances the detection efficacy of deceptive responses. By combining $Challenge$ and $Explain$ strategies, the interrogation process collects more detailed information about MLLMs' reasoning, which is crucial for accurately detecting overconfidence behaviors.
 
\noindent\textbf{Strategy Combination.} The overall prompt design references existing templates from relevant literature~\cite{ni2024retrieval}. In practical application scenarios, the system provides an expandable strategy combination space: the activation state of the punishment mechanism constitutes a binary choice, while at least one of the $Challenge$ and $Explain$ strategies must be activated, resulting in six combination modes (2×3). This modular design enables multi-dimensional exploration of the MLLMs' cognitive boundaries~\cite{kadavath2022,liu2023uncertainty} through strategy combinations and can be innovatively adjusted according to specific research needs.

\subsection{Auxiliary Expert LLM Assessment}
In this section, we template the previously collected information ($Q$, $A$, $C$, $R_{mllm}$) and input it into an expert LLM (llama3-instruct-8B, in our practice) for providing calibrated evaluation of the MLLMs' responses. The expert LLM's output includes the reassessed confidence $C_r$ adjusted based on the former interrogation process, as well as the reasoning $R_{exp}$ for $C_r$ that helps better understand the MLLMs' behavior. This information can be used to comprehensively evaluate the MLLMs' performance and identify areas where MLLMs may be prone to errors or inconsistencies. By comparing $C$ and $C_r$, MLLMs' introspective ability can be quantified, while the $R_{exp}$ generated by the expert LLM reveals the MLLMs' decision bias patterns, providing interpretable insights for model optimization.


\section{Experiments and Results }

\subsection{Experimental Setup}

\textbf{Datasets.} We extensively evaluate our proposed method on three public medical VQA datasets:

\begin{itemize}[label=\textbullet,itemsep=0pt,topsep=0pt,parsep=0pt]
    \item \textbf{Med-VQA}~\cite{hasan2018overview}: contains over 4,000 medical images and 50,000 question-answer pairs.
    \item \textbf{VQA-RAD}~\cite{lau2018dataset}: includes 315 radiological images annotated by clinicians and 3,515 question-answer pairs.
    \item \textbf{SLAKE}~\cite{liu2021slake}: a semantically annotated knowledge-enhanced medical VQA dataset with 642 images and 14,000 bilingual question-answer pairs.
\end{itemize}

Ultimately, we select 1,179 closed-ended questions from the test sets of the three datasets as a benchmark for evaluating MLLMs' capabilities and self-awareness.

\noindent\textbf{MLLM Backbones.} We test three MLLMs to evaluate their performance on medical VQA tasks under various calibration methods (including ours):

\begin{itemize}[label=\textbullet,itemsep=0pt,topsep=0pt,parsep=0pt]
    \item \textbf{LLaVA-1.5-Med-Mistral-7b}~\cite{li2024llavamed}: a multimodal large model specifically designed for the biomedical field, demonstrates excellent cross-modal understanding abilities.
    \item \textbf{LLaVA-NeXT-Mistral-7b}~\cite{llavanext2024}: an upgraded version of LLaVA-1.5~\cite{liu2024improved}, shows significant improvements in visual dialogue and reasoning capabilities.
    \item \textbf{Molmo-7b}~\cite{deitke2024molmo}: a general-purpose MLLM developed by Ai2, and its smaller parameter model outperforms models with 10 times more parameters.
    \item \textbf{MedVLM-R1}~\cite{pan2025}: a recent medical reasoning model that leverages reinforcement learning, specifically the GRPO algorithm, achieves strong results in medical VQA.
\end{itemize}

\noindent\textbf{Metrics.} We evaluate MLLM calibration using two core metrics. Expected Calibration Error \textbf{(ECE)} \cite{guo2017calibration} quantifies the deviation between model confidence and predictive accuracy by partitioning confidence scores into $M$ equally spaced bins $B_m$ ($m=1,\dots,M$). It calculates the weighted difference between average accuracy and confidence within each bin. Due to its intuitiveness, it is also widely regarded as the primary metric for assessing the calibration degree of models. The metric is formulated as:
\begin{equation}
\text{ECE} = \sum_{m=1}^{M} \frac{|B_m|}{N} |\text{acc}(B_m) - \text{conf}(B_m)|
\end{equation}
Area Under Receiver Operating Characteristic Curve \textbf{(AUROC)}~\cite{bradley1997roc} assesses confidence scores' diagnostic capability in distinguishing correct predictions via ROC curve integration. The metric is formulated as:
\begin{equation}
\text{AUROC} = \frac{1}{|C^+||C^-|} \sum_{C_i \in C^+} \sum_{C_j \in C^-} \mathbb{I}(C_i > C_j)
\end{equation}
where $C_i$ and $C_j$ denote confidence scores for correct and incorrect predictions respectively, and $\mathbb{I}(\cdot)$ is an indicator function returning $1$ when $C_i > C_j$ and $0$ otherwise.


\begin{table}
\caption{Performance comparison of different calibration methods across MLLMs and medical VQA datasets (Values are converted to percentages, and the optimal and suboptimal results in each column are highlighted in bold and underlined, respectively. The same applies to Table \ref{tab:ablation}.)}\label{tab:results}
\begin{tabular}{|l|l|cc|cc|cc|cc|}
\hline
\multirow{2}{*}{Model} & \multirow{2}{*}{Method} & \multicolumn{2}{c|}{med-vqa} & \multicolumn{2}{c|}{vqa-rad} & \multicolumn{2}{c|}{slake} & \multicolumn{2}{c|}{Avg.} \\
\cline{3-10}
& & ECE↓ & AUC↑ & ECE↓ & AUC↑ & ECE↓ & AUC↑ & ECE↓ & AUC↑ \\
\hline
\multirow{5}{*}{Llava-1.5-med-7B}
& Vanilla & 47.71& 48.46& 39.44& 50.63& 45.70& 49.36& 44.28& 49.48\\
& Punish & 46.68& 48.47& \underline{38.31}& \underline{51.30}& 42.43& 49.55& \underline{42.47}& 49.77\\
& Top-K & \underline{43.85}& \underline{50.00}& 42.41& 50.00& \underline{41.69}& \underline{50.00}& 42.65& \underline{50.00}\\
& \textbf{Ours} & \textbf{27.89}& \textbf{55.70}& \textbf{21.77}& \textbf{56.80}& \textbf{29.00}& \textbf{52.70}& \textbf{26.22}& \textbf{55.07}\\
\hline
\multirow{5}{*}{Llava-NeXT-7B}
& Vanilla & 25.94& 57.01& 35.16& 54.56& 37.00& \textbf{57.48}& \underline{32.70}& 56.35\\
& Punish & \underline{25.58}& \underline{58.28}& 36.37& \underline{55.85}& 37.04& \underline{56.65}& 33.00& \underline{56.93}\\
& Top-K & 27.80& 47.43& \underline{33.69}& 54.03& \underline{36.88}& 53.25& 32.79& 51.57\\
& \textbf{Ours} & \textbf{13.80}& \textbf{58.44}& \textbf{20.90}& \textbf{56.88}& \textbf{19.52}& 55.73& \textbf{18.07}& \textbf{57.02}\\
\hline
\multirow{5}{*}{Molmo-7B}
& Vanilla & 25.08& 51.45& \underline{30.28}& 49.87& \underline{26.87}& \textbf{54.45}& \underline{27.68}& \underline{51.92}\\
& Punish & \underline{23.54}& \underline{51.69}& 32.47& 46.98& 29.35& 50.19& 30.91& 49.62\\
& Top-K & 39.28& 45.46& 31.04& \textbf{56.47}& 48.71& 39.48& 39.68& 47.14\\
& \textbf{Ours} & \textbf{14.52}& \textbf{62.20}& \textbf{24.52}& \underline{52.39}& \textbf{20.22}& \underline{51.25}& \textbf{19.75}& \textbf{55.28}\\
\hline
\multirow{5}{*}{MedVLM-R1}
& Vanilla & 20.82 & 49.15 & \underline{25.59} & 54.95 & 30.46 & 44.81 & 25.62 & 49.64 \\
& Punish & 24.58 & 49.71 & 27.63 & 57.59 & 33.44 & 42.38 & 28.55 & 49.89 \\
& Top-K & \underline{13.24} & \underline{62.51} & 28.91 & \textbf{59.29} & \underline{16.76} & \textbf{60.59} & \underline{19.64} & \textbf{60.80} \\
& \textbf{Ours} & \textbf{12.52} & \textbf{63.32} & \textbf{16.85} & \underline{59.03} & \textbf{14.06} & \underline{55.62} & \textbf{14.48} & \underline{59.32} \\
\hline
\end{tabular}
\end{table}

\noindent\textbf{Baselines.} We test the performance of three LLM confidence calibration methods:

\begin{itemize}[label=\textbullet,itemsep=0pt,topsep=0pt,parsep=0pt]
\item   \textbf{Vanilla}~\cite{xiong2023uncertainty}: directly extracts the model's confidence in the verbal answer. Its advantage is that it provides a basic confidence indicator without additional computation.

\item   \textbf{Punish}~\cite{ni2024retrieval}: adds a prompt "You will be punished if the answer is wrong but you answer it with high confidence" to encourage the model to be cautious in answering.

\item   \textbf{Top-K}~\cite{tian2023calibration}: requires the model to provide $k$ best guesses $\{G_1, \ldots, G_k\}$ with corresponding probabilities $\{P_1, \ldots, P_k\}$, eliciting verbalized likelihoods. For fair comparison across methods, we apply computational normalization by scaling the maximum probability to confidence through linear transformation:
\begin{equation}
\text{Confidence} = \max\{P_1, \ldots, P_k\} \times 10.    
\end{equation}
\end{itemize}

\subsection{Main Results}

As shown in Table \ref{tab:results}, we evaluate four confidence calibration methods for the medical VQA task through comparative experiments. Notably, our proposed method outperforms the baseline methods across all the medical VQA datasets and MLLMs, reducing the average ECE of LLaVA-1.5-med-7B from 44.28\% to 26.22\% (a 40.8\% reduction) and increasing AUROC from 49.48\% to 55.07\%. This demonstrates significant improvement in both calibration and the model's ability to distinguish between correct and incorrect answers. Moreover, our findings show that domain-specific characteristics significantly affect calibration performance. General-domain MLLMs, like Molmo-7B, exhibit low ECE across calibration methods, with its vanilla method achieving an average ECE of 27.68\%. In contrast, medical-domain MLLMs, such as LLaVA-1.5-med-7B, show persistent overconfidence, with an average ECE of 44.28\%, much higher than general-domain MLLMs. This suggests that supervised fine-tuning (SFT) on domain-specific data may impair confidence calibration~\cite{ren2024sft}.

\begin{table}
\caption{Performance comparison of different strategy combinations.}\label{tab:ablation}
\fontsize{8}{9.6}\selectfont
\begin{tabular}{|l|c|c|c|c|c|c|c|c|c|}
\hline
\multirow{2}{*}{Method} & \multicolumn{3}{c|}{Llava-1.5-med-7B} & \multicolumn{3}{c|}{Llava-NeXT-7B} & \multicolumn{3}{c|}{Molmo-7B} \\
\cline{2-10}
& \makebox[1.045cm][c]{med-vqa} & \makebox[1.045cm][c]{vqa-rad} & \makebox[1.045cm][c]{slake}  & \makebox[1.045cm][c]{med-vqa} & \makebox[1.045cm][c]{vqa-rad} & \makebox[1.045cm][c]{slake}  & \makebox[1.045cm][c]{med-vqa} & \makebox[1.045cm][c]{vqa-rad} & \makebox[1.045cm][c]{slake} \\
\hline
Ch. & 32.36 & 21.52 & \underline{27.47} & 14.04 & \textbf{19.21} & \underline{20.31} & \underline{17.00} & \underline{25.16} & \underline{22.09} \\
Exp. & 34.46 & 21.53 & 35.93 & 16.92 & 26.49 & 26.15 & 27.72 & 28.03 & 29.57 \\
Ch.+Exp. & 30.51 & \underline{20.82} & 30.74 & 14.42 & 21.04 & 22.19 & 21.80 & 25.95 & 25.96 \\
Pu+Exp. & 34.00 & 22.29 & 29.72 & 14.89 & 25.32 & 24.33 & 26.58 & 27.06 & 27.07 \\
\textbf{Pu.+Ch.} & \textbf{27.89} & 21.77 & 29.00 & \textbf{13.80} & \underline{20.90} & \textbf{19.52} & \textbf{14.52} & \textbf{24.52} & \textbf{20.22} \\
Pu+Ch+Exp. & \underline{29.28} & \textbf{17.90} & \textbf{25.77} & \underline{13.88} & 22.37 & 21.54 & 20.52 & 25.45 & 23.12 \\
\hline
\end{tabular}
\end{table}

\subsection{Ablation Study}

Ablation experiments reveal that strategy combinations significantly impact the calibration performance of medical visual question answering models, measured by ECE. The $Punish$ only approach (Pu.) was excluded from the table as it omitted the model's rebuttal phase, making the expert model's calibrated confidence scores unreliable. The $Punish$ and $Challenge$ (Pu.+Ch.) combination proved most effective on the Llava-1.5-med-7B model, with an ECE of 27.89\%, and also demonstrated optimal performance in cross-model scenarios, reducing Molmo-7B's ECE by 14.6\%. However, strategy effectiveness is not tied to complexity, as combining all three methods yielded suboptimal results. 

Furthermore, the experiments show cross-model heterogeneity: the single $Challenge$ (Ch.) strategy was superior for the Llava-NeXT-7B model on the vqa-rad dataset, while Pu.+Ch. performed best on the slake dataset for multiple models. This variation stems from pre-training differences, where overconfident models like Llava-Med benefit from multi-strategy intervention, while more capable models like Molmo-7B perform better with simpler strategies.
\begin{figure}
\includegraphics[width=\textwidth]{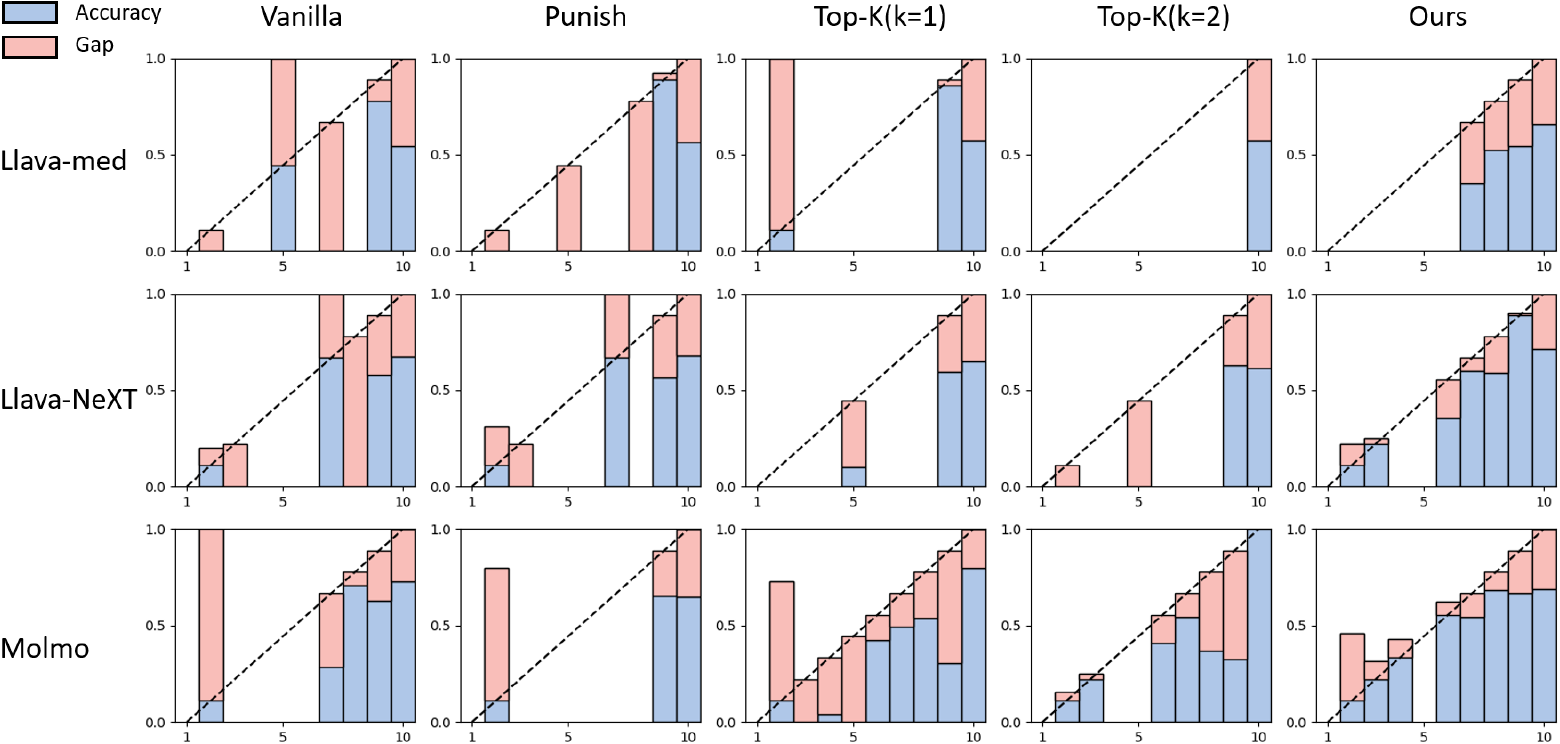}
\caption{Visualization analysis of the confidence (x-axis) vs. accuracy (y-axis) calibration comparison across different baselines (including ours), with all datasets aggregated.} \label{fig2}
\end{figure}
\subsection{Visualization and Analysis}
Figure \ref{fig2} demonstrates a comparison of the calibration effects between different calibration methods across three datasets, with confidence (x-axis) and accuracy (y-axis). The visual analysis reveals significant differences in calibration characteristics under various models (Llava-med, Llava-NeXT, and Molmo) for each calibration method. Specifically, the Llava-med model exhibits a clear calibration underperformance under the Vanilla method, showing a significant discrepancy between confidence and accuracy. In contrast, our MS-FBI method substantially reduces the gap between confidence and accuracy, showing superior calibration performance. Similar trends are observed in the Llava-NeXT and Molmo models, confirming the method's advantage in cross-model generalization. Overall, the method proposed in this paper demonstrates the best calibration effects across all models and datasets, effectively enhancing the consistency between model confidence and accuracy.

\section{Discussion and Conclusion}
This study is the first to combine multi-strategy fusion-based interrogation with expert LLM evaluation framework to explore the calibration of medical MLLMs, achieving significant progress. The experiments show that the proposed method effectively improves the alignment between model confidence and prediction accuracy, and reveals MLLM's decision-making patterns in complex medical scenarios through a multi-round interrogation mechanism. Ablation experiments confirm the contributions of optimization strategies and expert LLMs to the calibration effect, providing support for enhancing model reliability. 

Although the generalizability of the method has been preliminarily validated, more refined calibration strategies may be needed in the case of general LLMs. The study also highlights key future research directions, such as domain-specific calibration for medical models, unification of expert evaluation standards, and achieving efficient calibration without increasing computational costs. Addressing these issues will advance MLLM calibration technologies and provide more reliable intelligent tools for clinical decision support systems.

\begin{credits}
\subsubsection{\ackname} This work was supported by the National Natural Science Foundation of China under Grant 42394060 and 42394064 and the Earth System Big Data Platform of the School of Earth Sciences, Zhejiang University.

\subsubsection{\discintname}
The authors have no competing interests to declare.
\end{credits}
%
%
%
\bibliographystyle{splncs04}
\bibliography{Paper-1840}

\end{document}